\documentclass{article}
\usepackage{spconf,amsmath,graphicx,hyperref,booktabs}
\title{BEYOND THE DOMINANT LABEL: RECOVERING COEXISTING LESION EVIDENCE WITH MULTIPLE INSTANCE LEARNING}

\name{Lubin Gan$^{1}$ \quad Jing Zhang$^{2,*}$ \quad Xinming Zhang$^{1}$ \quad
Chengpeng Xu$^{1}$ \quad Siying Wu$^{2}$ \quad Xiaoyan Sun$^{1,2,*}$\thanks{*Corresponding authors: Jing Zhang and Xiaoyan Sun.}}
\address{$^{1}$USTC, Anhui, China \\
$^{2}$Anhui Province Key Laboratory of Biomedical Imaging and Intelligent Processing \\
Institute of Artificial Intelligence, Hefei Comprehensive National Science Center, Anhui, China}

\begin{document}
\ninept
\maketitle

\begin{abstract}
Whole-slide images may contain coexisting lesions, yet weak supervision typically
provides only the most severe diagnosis. We formulate this setting as
dominant-label multiple instance learning, where the observed label is the
maximum-severity element of a latent lesion set. Standard multiclass MIL treats
it as a complete one-hot description and therefore suppresses lower-severity
evidence whenever it co-occurs with a dominant lesion. We propose DomSet-MIL, a
label-aware framework that aligns the annotation mechanism across bag and
instance levels. Dominant-Label Censored Likelihood models the reported diagnosis
without declaring unreported lesions absent. Severity-Ordered Residual Routing
assigns patch evidence from high to low severity and preserves residual support
for secondary findings, while top-$q$ pooling links local routing scores to bag
prediction. Experiments across natural and controlled settings show that
DomSet-MIL achieves state-of-the-art performance in dominant-label prediction
while more effectively recovering latent coexisting-lesion evidence. These
findings demonstrate that modeling the dominant-label mechanism can
improve diagnostic prediction without sacrificing clinically meaningful
secondary findings.
\end{abstract}


\section{Introduction}
\label{sec:intro}

Whole-slide images (WSIs) provide a comprehensive view of tissue morphology, but
their gigapixel resolution makes dense region-level annotation impractical.
Weakly supervised and multiscale systems address this challenge without exhaustive
region annotations~\cite{campanella2019clinical,hashimoto2020multiscale};
multiple instance learning (MIL)~\cite{dietterich1997mil} represents a WSI as
patch embeddings learned from one slide-level diagnosis.
Complementary work on zero-shot subtyping uses fine-grained patch--text
alignment to connect local morphology with pathology-aware descriptions,
as in FG-PAN~\cite{gan2026enhancing}.

Attention pooling~\cite{ilse2018attention} and pathology-specific methods such as
CLAM~\cite{lu2021clam} identify discriminative instances, while Transformer
aggregators model their dependencies~\cite{shao2021transmil}.
SemaMIL combines semantic patch reordering with retrieval-guided state space
modeling~\cite{gan2026semamil}. Despite their
different aggregation mechanisms, they usually assume the observed bag label is a mutually exclusive
class that fully describes the slide, an assumption that may not match multiclass
pathology labels.

Preserving discriminative information is also studied in graph representation
learning. GraphRefiner addresses generic-token dominance
in graph--text alignment~\cite{zhang2026mitigating}, whereas H4G studies
fine-grained structural information in hyperbolic graph
representations~\cite{zhang2025h4g}.

\begin{figure}[!t]
    \centering
    \includegraphics[width=\columnwidth]{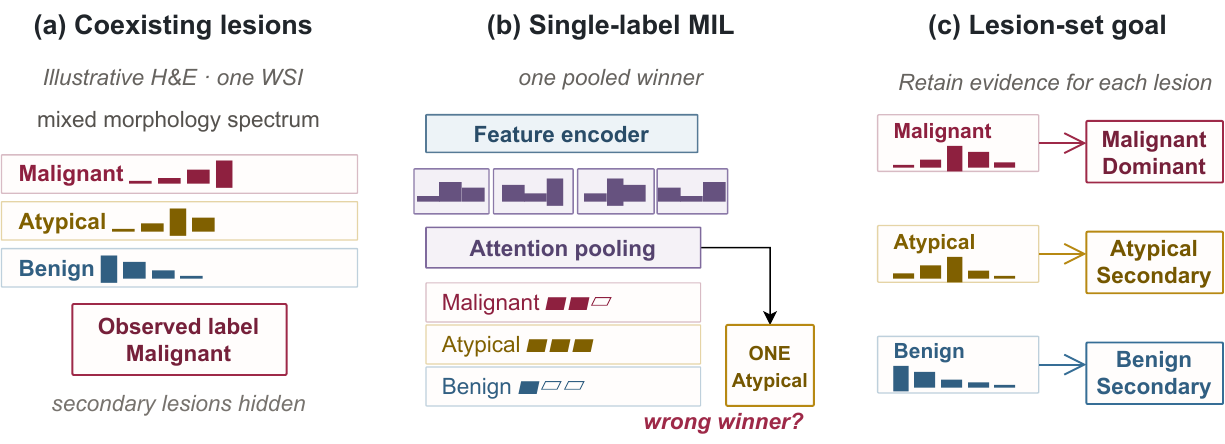}
    \caption{Motivation for lesion-set learning under dominant-label supervision.
    Coexisting lesions are summarized by a single dominant label in (a);
    single-label MIL produces one pooled winner in (b), whereas lesion-set
    learning retains evidence for each lesion in (c).}
    \label{fig:dominant_label_motivation}
\end{figure}

\begin{figure*}[!t]
    \centering
    \includegraphics[width=\textwidth]{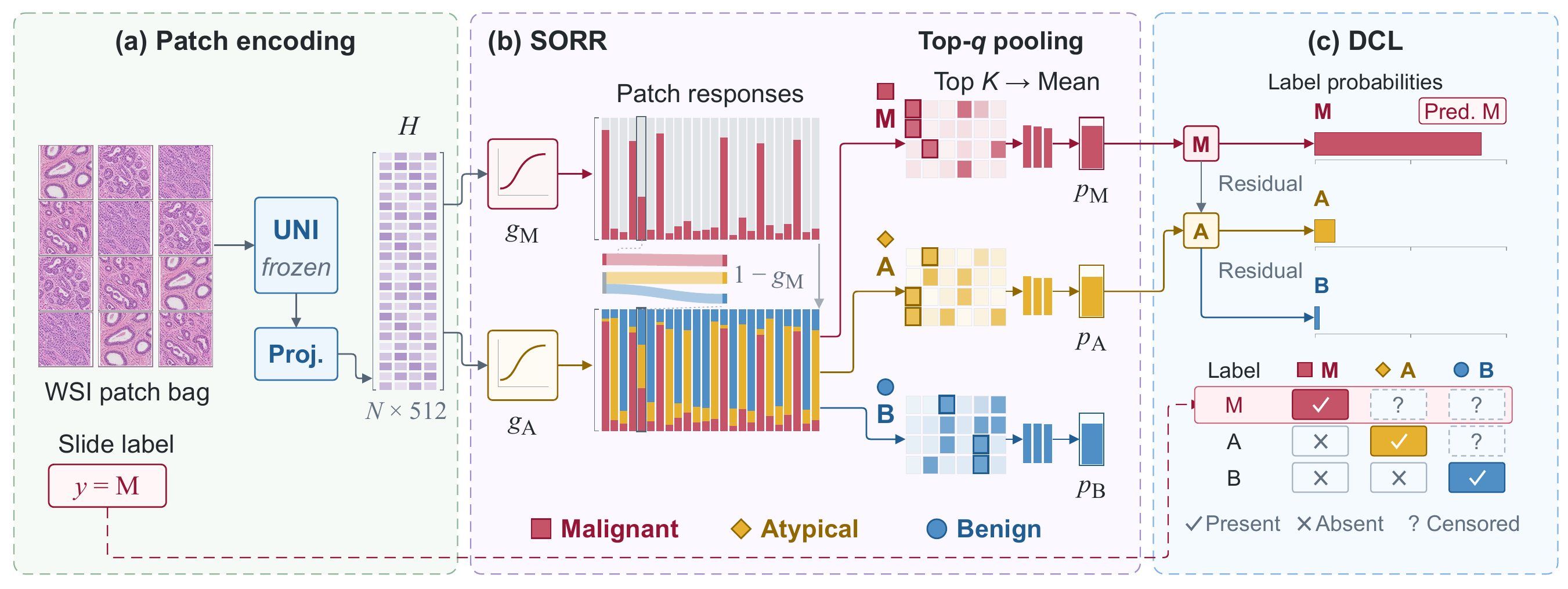}
    \caption{Overview of DomSet-MIL: SORR routes patch evidence in severity order, and DCL predicts the dominant label.}
    \label{fig:framework}
\end{figure*}

This mismatch arises when a heterogeneous specimen is summarized by its dominant
finding. A pathology WSI can contain regions with different diagnostic significance~\cite{mercan2018miml,brancati2022bracs},
yet weak supervision may expose only the most severe lesion. The annotation
therefore records the maximum severity of a latent lesion set, rather than the
set itself: a slide containing malignant and atypical tissue is labeled
malignant, so standard softmax treats atypical as an incorrect bag label. Across
such cases, unreported secondary lesions become structural false negatives;
lower-severity labels are censored whenever they co-occur with a more severe
finding, and changing the aggregator cannot remove supervision introduced by
this annotation rule.

Recent work incorporates diagnostic severity into multiclass MIL. PAMS organizes
classes hierarchically, penalizes clinically costly underestimation, and mixes
features across priorities~\cite{hong2026pams}, but still targets only the final
diagnosis, so lower-severity evidence need not remain positive. Multi-instance
multi-label methods predict coexisting breast lesions but assume multiple
diagnoses are observed during training~\cite{mercan2018miml}. Learning from a
single positive label relaxes that assumption by leaving other labels unknown~\cite{cole2021singlepositive}, yet generic
missing-label objectives neither explain why one label was selected nor link
bag-level presence to localized evidence. This gap calls for explicit modeling of the dominant-label annotation mechanism.

We propose \emph{DomSet-MIL}, a label-aware framework that predicts the dominant diagnosis while preserving evidence for hidden lesions. At the bag level, Dominant-Label Censored Likelihood (DCL) converts a malignant presence gate and a residual atypical gate into three observed class probabilities via the annotation's maximum-severity rule; a malignant label supervises malignant presence without declaring atypical absence. At the instance level, Severity-Ordered Residual Routing (SORR) applies the rule in descending severity: a malignant gate claims matching patches, an atypical gate processes residual evidence, and unclaimed mass forms the benign remainder. Top-$q$ pooling connects these severity-specific patch assignments to DCL's censored bag scores. The two components therefore impose the same severity-ordered factorization at complementary resolutions.

Our contributions are threefold: (1) We identify lower-severity lesions as
structurally censored under dominant-label supervision, explaining their
suppression as negative classes. (2) We introduce DomSet-MIL, coupling DCL with
SORR to align latent evidence at bag and patch levels. (3) We evaluate
dominant-label prediction and latent-lesion recovery on BRACS~\cite{brancati2022bracs},
with test-time ROI annotations and on controlled EBRAINS/TCGA mixed-bag benchmarks.

\section{Method}
\label{sec:method}

\subsection{Dominant-label formulation}

Let $\mathcal{D}=\{(\mathcal{B}_i,y_i)\}_{i=1}^{I}$ denote the training set of
$I$ WSIs. Slide $i$ is represented by a bag
$\mathcal{B}_i=\{\mathbf{x}_{in}\}_{n=1}^{N_i}$ of $N_i$ patch embeddings
extracted by a fixed encoder. We consider the ordered diagnosis set
$\mathcal{C}=\{\mathrm{B},\mathrm{A},\mathrm{M}\}$, where
$\mathrm{B}\prec\mathrm{A}\prec\mathrm{M}$ denote benign, atypical, and
malignant findings, respectively. Each slide contains an unobserved, nonempty
lesion set $S_i\subseteq\mathcal{C}$, whereas training provides only its most
severe element:
\begin{equation}
    y_i=\max_{\prec} S_i.
    \label{eq:dominant_rule}
\end{equation}
Consequently, $y_i=\mathrm{M}$ confirms malignant tissue but leaves atypical
presence unknown. In contrast, $y_i=\mathrm{A}$ confirms atypical tissue and
rules out malignancy, while $y_i=\mathrm{B}$ rules out both higher-severity
findings. DomSet-MIL encodes these three supervision states instead of treating
$y_i$ as a complete one-hot description of the slide. The complete framework is
illustrated in Fig.~\ref{fig:framework}.

\subsection{Severity-ordered residual routing}

Model specialization has been studied through geometry-aware dense-to-expert
reconstruction of
vision--language models~\cite{zhang2026geometry}; the AsyMoE preprint discussed
modality-specific experts but has been withdrawn~\cite{zhang2025asymoe}.
SORR assigns severity-specific patch mass using ordered gates.

We first map each fixed embedding to a trainable representation,
\begin{equation}
\begin{aligned}
    \mathbf{h}_{in}
        &=\operatorname{ReLU}(\mathbf{W}\mathbf{x}_{in}+\mathbf{b}),\\
    g_{in,c}
        &=\sigma(\mathbf{w}_c^\top\mathbf{h}_{in}+a_c),
          &&c\in\{\mathrm{M},\mathrm{A}\},
\end{aligned}
    \label{eq:patch_gates}
\end{equation}
where $\mathbf{W}$ and $\mathbf{b}$ form a shared projection,
$(\mathbf{w}_c,a_c)$ parameterize a class gate, and $\sigma$ is the logistic
sigmoid. SORR converts the two gate values into severity-specific routing masses
by applying the malignant gate first and restricting the atypical contribution
to its residual:
\begin{align}
    \pi_{in,\mathrm{M}} &= g_{in,\mathrm{M}}, \nonumber\\
    \pi_{in,\mathrm{A}} &= (1-g_{in,\mathrm{M}})g_{in,\mathrm{A}}, \label{eq:sorr}\\
    \pi_{in,\mathrm{B}} &= (1-g_{in,\mathrm{M}})(1-g_{in,\mathrm{A}}). \nonumber
\end{align}
Thus, $\sum_{c\in\mathcal{C}}\pi_{in,c}=1$ for every patch, and benign evidence
is the parameter-free remainder rather than a third competing gate. The local
residual limits atypical support to the mass left by the malignant gate;
it does not suppress atypical evidence elsewhere.

G$^2$rammar explicitly encodes structural and
semantic roles for text-attributed graphs~\cite{zheng2026g2rammar}.
An earlier preprint on structural-role inference with LLMs has been
withdrawn~\cite{zhang2025empowering}.

To obtain bag scores without diluting sparse lesions or saturating a noisy-OR
operator, we retain a fixed fraction of the strongest responses. Let
$q\in(0,1]$, $K_i=\max\{1,\lceil qN_i\rceil\}$, and
$\mathcal{T}_{i,c}$ be the indices of the $K_i$ largest values in
$\{\pi_{in,c}\}_{n=1}^{N_i}$. We define a class-specific evidence score for
every route as
\begin{equation}
    p_{i,c}=\frac{1}{K_i}\sum_{n\in\mathcal{T}_{i,c}}\pi_{in,c},
    \qquad c\in\mathcal{C}.
    \label{eq:topq}
\end{equation}
Because $K_i$ scales with $N_i$, the same $q$ has a consistent coverage meaning
across slides with different tissue areas. DCL uses $p_{i,\mathrm{M}}$ and
$p_{i,\mathrm{A}}$ to model the observed label, whereas $p_{i,\mathrm{B}}$
supports routing.

Higher-order grouping is another possible topic for future aggregation
research. HyperAgent investigates hypergraph communication
topologies~\cite{zhang2026hyperagent}, and EvoHyper represents collaboration
groups and shared memory using evolving hyperedges~\cite{zhang2026evohyper}.

\begin{table*}[!t]
    \centering
    \caption{Dominant-label classification on BRACS and controlled mixed-bag
    tasks. Best results are bold.}
    \label{tab:main_results}
    \footnotesize
    \setlength{\tabcolsep}{2.2pt}
    \renewcommand{\arraystretch}{1.02}
    \begin{tabular*}{\textwidth}{@{\extracolsep{\fill}}lccccccccc@{}}
        \toprule
        & \multicolumn{3}{c}{BRACS}
        & \multicolumn{2}{c}{EBRAINS-Mix}
        & \multicolumn{2}{c}{NSCLC-Mix}
        & \multicolumn{2}{c}{RCC-Mix} \\
        \cmidrule(lr){2-4}\cmidrule(lr){5-6}\cmidrule(lr){7-8}\cmidrule(lr){9-10}
        Method & M-AUC & BAcc & UER$\downarrow$
               & M-AUC & BAcc & M-AUC & BAcc & M-AUC & BAcc \\
        \midrule
        \multicolumn{10}{l}{\textit{Published MIL methods}} \\
        ABMIL                 & 0.881 & 0.643 & 0.118 & 0.897 & 0.804 & 0.936 & 0.867 & 0.892 & 0.781 \\
        CLAM-SB               & 0.887 & 0.652 & 0.110 & 0.903 & 0.814 & 0.942 & 0.876 & 0.899 & 0.793 \\
        DSMIL                 & 0.894 & 0.661 & 0.101 & 0.914 & 0.827 & 0.950 & 0.887 & 0.908 & 0.806 \\
        TransMIL              & 0.901 & 0.668 & 0.096 & 0.909 & 0.820 & 0.947 & 0.883 & 0.906 & 0.804 \\
        DTFD-MIL (AFS)        & 0.899 & 0.666 & 0.098 & 0.917 & 0.832 & 0.952 & 0.890 & 0.912 & 0.812 \\
        ASMIL (ABMIL)         & 0.906 & 0.675 & 0.092 & 0.910 & 0.821 & 0.946 & 0.884 & 0.905 & 0.800 \\
        MAMMOTH (ABMIL)       & 0.912 & 0.684 & 0.086 & 0.918 & 0.833 & 0.953 & 0.891 & 0.915 & 0.816 \\
        SemaMIL               & 0.907 & 0.678 & 0.089 & 0.915 & 0.830 & 0.951 & 0.889 & 0.913 & 0.814 \\
        PAMS (DTFD-AFS)       & 0.910 & 0.681 & 0.073 & 0.916 & 0.831 & 0.952 & 0.890 & 0.914 & 0.814 \\
        \midrule
        \multicolumn{10}{l}{\textit{Matched label-model controls}} \\
        Flat-CE               & 0.883 & 0.646 & 0.119 & 0.891 & 0.795 & 0.931 & 0.861 & 0.884 & 0.772 \\
        CR-Ordinal            & 0.899 & 0.667 & 0.083 & 0.910 & 0.819 & 0.946 & 0.883 & 0.906 & 0.801 \\
        WAN                   & 0.892 & 0.658 & 0.102 & 0.906 & 0.814 & 0.944 & 0.878 & 0.901 & 0.791 \\
        DomSet-MIL            & \textbf{0.916} & \textbf{0.690} & \textbf{0.071}
                              & \textbf{0.922} & \textbf{0.842}
                              & \textbf{0.958} & \textbf{0.900}
                              & \textbf{0.920} & \textbf{0.826} \\
        \bottomrule
    \end{tabular*}
\end{table*}

\subsection{Dominant-label censored likelihood}

DCL interprets $p_{i,\mathrm{M}}$ as the malignant gate at bag level and
$p_{i,\mathrm{A}}$ as the atypical gate on the residual branch, conditional on
malignant evidence not being dominant. Applying the annotation rule in
Eq.~\eqref{eq:dominant_rule} gives a normalized distribution over observed
labels, where $P_i(c)=\Pr(y_i=c\mid\mathcal{B}_i)$:
\begin{align}
    P_i(\mathrm{M}) &= p_{i,\mathrm{M}}, \nonumber\\
    P_i(\mathrm{A}) &= (1-p_{i,\mathrm{M}})p_{i,\mathrm{A}}, \label{eq:dcl_prob}\\
    P_i(\mathrm{B}) &= (1-p_{i,\mathrm{M}})(1-p_{i,\mathrm{A}}). \nonumber
\end{align}
The outer factor $(1-p_{i,\mathrm{M}})$ enforces bag-level dominance,
whereas Eq.~\eqref{eq:sorr} resolves within-patch competition. This
continuation-ratio form~\cite{tutz2022ordinal} requires no independence assumption.
\begin{equation}
\ell_i=
\begin{cases}
 -\log p_{i,\mathrm{M}},
     & y_i=\mathrm{M},\\
 -\log\!\left[(1-p_{i,\mathrm{M}})p_{i,\mathrm{A}}\right],
     & y_i=\mathrm{A},\\
 -\log\!\left[(1-p_{i,\mathrm{M}})(1-p_{i,\mathrm{A}})\right],
     & y_i=\mathrm{B},
\end{cases}
\label{eq:dcl_loss}
\end{equation}
The model is trained with
$\mathcal{L}_{\mathrm{DCL}}=I^{-1}\sum_{i=1}^{I}\ell_i$.
For a malignant slide, the loss is independent of $p_{i,\mathrm{A}}$; atypical
evidence is therefore censored rather than penalized as a false positive. Its
score on such slides is transferred through gates shared with atypical and
benign training slides and should be interpreted as latent evidence, not an
identifiable multi-label target. At inference, the dominant diagnosis is
$\hat y_i=\arg\max_{c\in\mathcal{C}}P_i(c)$, while $p_{i,c}$ and
$\pi_{in,c}$ provide bag- and patch-level evidence. No ROI annotation is used to
learn these quantities.
For a two-level ordered task, the construction reduces to one high-priority
gate, with $P_i(\mathrm{H})=p_{i,\mathrm{H}}$ and
$P_i(\mathrm{L})=1-p_{i,\mathrm{H}}$.

\section{Experiments}
\label{sec:experiments}

\begin{table*}[!t]
    \centering
    \caption{Latent-evidence recovery on controlled mixed bags and BRACS ROIs.
    Dashes mark unavailable ROI metrics; best results are bold.}
    \label{tab:evidence_recovery}
    \footnotesize
    \setlength{\tabcolsep}{2.2pt}
    \renewcommand{\arraystretch}{1.02}
    \begin{tabular*}{\textwidth}{@{\extracolsep{\fill}}lcccccccc@{}}
        \toprule
        & \multicolumn{6}{c}{\textit{Bag level}}
        & \multicolumn{2}{c}{\textit{ROI level}} \\
        \cmidrule(lr){2-7}\cmidrule(lr){8-9}
        & \multicolumn{2}{c}{EBRAINS-Mix}
        & \multicolumn{2}{c}{NSCLC-Mix}
        & \multicolumn{2}{c}{RCC-Mix}
        & \multicolumn{2}{c}{BRACS} \\
        \cmidrule(lr){2-3}\cmidrule(lr){4-5}\cmidrule(lr){6-7}\cmidrule(lr){8-9}
        Method & Set-mAP & Hidden-R & Set-mAP & Hidden-R & Set-mAP & Hidden-R
               & ROI-mAP & Sec-ROI-R \\
        \midrule
        ABMIL                 & 0.618 & 0.286 & 0.661 & 0.318 & 0.592 & 0.251 & -- & -- \\
        DSMIL                 & 0.653 & 0.349 & 0.702 & 0.386 & 0.636 & 0.317 & 0.556 & 0.392 \\
        DTFD-MIL (AFS)        & 0.671 & 0.371 & 0.718 & 0.401 & 0.654 & 0.337 & -- & -- \\
        SemaMIL               & 0.664 & 0.362 & 0.712 & 0.396 & 0.649 & 0.331 & -- & -- \\
        PAMS (DTFD-AFS)       & 0.642 & 0.322 & 0.696 & 0.365 & 0.628 & 0.294 & -- & -- \\
        \midrule
        Flat-CE               & 0.604 & 0.244 & 0.647 & 0.282 & 0.581 & 0.219 & 0.523 & 0.347 \\
        CR-Ordinal            & 0.626 & 0.301 & 0.677 & 0.346 & 0.606 & 0.274 & -- & -- \\
        WAN                   & 0.713 & 0.451 & 0.754 & 0.486 & 0.696 & 0.426 & 0.594 & 0.451 \\
        DomSet-MIL            & \textbf{0.748} & \textbf{0.547}
                              & \textbf{0.789} & \textbf{0.582}
                              & \textbf{0.734} & \textbf{0.515}
                              & \textbf{0.625} & \textbf{0.500} \\
        \bottomrule
    \end{tabular*}
\end{table*}

\subsection{Datasets and evaluation protocol}

Spatial--semantic consistency is studied in multi-contrast MRI super-resolution
by SSCM~\cite{wu2026sscm}. Our evaluation focuses on histopathology and assesses
both slide-level predictions and localized lesion evidence.

\textbf{BRACS.}
BRACS contains 547 breast WSIs from 189 patients and provides both slide labels
and labeled regions of interest (ROIs)~\cite{brancati2022bracs}. We use its
official patient-disjoint split of 395/67/85 WSIs for training, validation, and
testing. Following the official coarse task, normal, pathological benign, and
usual ductal hyperplasia are grouped as benign; flat epithelial atypia and
atypical ductal hyperplasia as atypical; and ductal carcinoma in situ and
invasive carcinoma as malignant. This mapping gives the order
$\mathrm{B}\prec\mathrm{A}\prec\mathrm{M}$ used in Sec.~\ref{sec:method}.
Training and model selection use only the coarse WSI label. ROI labels are
reserved for the test-time evidence analysis below.

\textbf{Controlled mixed-bag tasks.}
We complement the natural BRACS setting with three controlled tasks for which
the latent contents of each bag are known by construction. From
EBRAINS~\cite{roetzer2022ebrains}, we select meningothelial, atypical, and
anaplastic meningioma and map their WHO-2016 grades~\cite{louis2016who} to the
low, middle, and high slots of DomSet-MIL. From TCGA~\cite{tcga2013pancancer},
we use LUAD versus LUSC~\cite{tcga2014luad,tcga2012lusc} for NSCLC subtyping and
KICH, KIRC, and KIRP~\cite{davis2014kich,tcga2013kirc,tcga2016kirp} for RCC
subtyping.
These TCGA subtypes have no asserted clinical severity order; their order is a
benchmark device used solely to impose a known censoring mechanism. We therefore
average NSCLC results over both class orders and RCC results over all six class
orders.

Patients are partitioned before synthesis. We use five patient-level folds for
EBRAINS and site-stratified patient-level folds for TCGA, with 10\% of the
training patients in each fold held out for validation. For a task with class
set $\mathcal{C}_d$, we sample the nonempty latent sets
$S_i\subseteq\mathcal{C}_d$ approximately uniformly. A bag with
$r=|S_i|$ draws one source WSI per class and an equal share of 2,048 patch
embeddings from every source. Sampling is without replacement when a slide
contains enough tissue patches and with replacement only to fill a shortfall.
Each epoch contains as many newly sampled bags as source WSIs in its training
fold; validation and test mixtures are fixed by seed and matched in number to
their respective source pools. The observed training label is
$y_i=\max_{\prec}S_i$, while $S_i$ remains hidden during model fitting. Thus,
all methods receive exactly the same dominant-label training supervision.

\subsection{Implementation details and baselines}

Tissue is segmented by Otsu thresholding~\cite{otsu1979threshold} and tiled into
non-overlapping
$224\times224$ patches at $20\times$ magnification. A frozen UNI
ViT-L/16~\cite{chen2024uni} maps every patch to a 1,024-dimensional feature.
The projection in Eq.~\eqref{eq:patch_gates} has dimension 512. We use
AdamW~\cite{loshchilov2019adamw} with learning rate $10^{-4}$, weight decay $10^{-5}$,
and one bag per update for up to 100 epochs. Early stopping uses validation macro-AUC with
a patience of 15 epochs. The pooling fraction is selected from
$q\in\{0.01,0.05,0.10,0.20\}$ on the validation set. BRACS results are averaged
over five seeds. Controlled-task results are averaged over folds, five seeds,
and, for TCGA, the class orders.

Established baselines are ABMIL~\cite{ilse2018attention} and
CLAM-SB~\cite{lu2021clam}. We additionally evaluate
DSMIL~\cite{li2021dsmil}, TransMIL~\cite{shao2021transmil}, and the AFS variant
of DTFD-MIL~\cite{zhang2022dtfd}. Recent comparisons include attention
stabilization applied to ABMIL (ASMIL)~\cite{ye2026asmil}; the ABMIL
instantiation of the task-specific mixture-of-experts projection MAMMOTH
\cite{shao2026mammoth}; and the state-space-based
SemaMIL~\cite{gan2026semamil}. The severity-aware PAMS framework is instantiated with
DTFD-AFS~\cite{hong2026pams}. All methods use the same frozen UNI features, data
splits, and model-selection budget.

ReconMIL explored latent-space reconstruction and bi-stream Mamba for WSI
analysis~\cite{gan2026reconmil}; its arXiv record was subsequently withdrawn
because of evaluation-protocol issues.

Three matched controls isolate label modeling from aggregation. \emph{Flat-CE} uses the shared patch projection and top-$q$ readout, replacing SORR with competing patch logits and DCL with softmax cross-entropy. \emph{CR-Ordinal} applies a two-gate continuation-ratio head to a pooled bag representation, without instance-level residual routing. \emph{WAN} uses independent sigmoid patch heads and the same top-$q$ readout, training with Cole et al.'s~\cite{cole2021singlepositive} weak-assume-negative loss, weighting each unreported class by $\gamma=1/(|\mathcal{C}|-1)$.

We use macro one-vs-rest AUC (M-AUC), a threshold-free metric that weights
classes equally, as the primary classification endpoint; balanced accuracy
(BAcc) is a complementary operating-point measure. For BRACS, we additionally
report the severity underestimation rate (UER), the fraction of test slides with
$\hat y_i\prec y_i$; lower is better. UER is omitted for the TCGA tasks because
their priority order is synthetic. All results are averaged point estimates and
are interpreted descriptively rather than as statistical-significance claims.

\noindent\textbf{Dominant-label classification.}
Table~\ref{tab:main_results} summarizes dominant-label classification.
On BRACS, DomSet-MIL obtains 0.916 M-AUC, 0.690 BAcc, and 0.071 UER,
compared with the strongest respective baseline estimates of 0.912, 0.684,
and 0.073. It also leads on all controlled tasks, improving over the strongest
baseline by 0.004--0.005 M-AUC and 0.009--0.010 BAcc.

\subsection{Latent-evidence recovery}

Evidence localization should be distinguished from failure attribution or
multi-agent disagreement analysis. The GraphTracer deep-search preprint
\cite{zhang2025graphtracer} and the GraphGeo geo-localization preprint
\cite{zheng2025graphgeo} addressed these respective problems; both preprints
have been withdrawn.

On the controlled tasks, each model's bag-level class scores are compared with the held-out multi-hot indicator of $S_i$. Set mean average precision (Set-mAP) is the primary, threshold-free latent-set recovery metric, computed over all classes. Hidden-lesion recall (Hidden-R) is a complementary, threshold-dependent diagnostic computed over pairs $(i,c)$ for which $c\in S_i$ and $c\prec y_i$; singleton bags do not enter its denominator. For Hidden-R only, each method and class uses a threshold selected on the validation set by maximizing F1 against the retained latent sets. This evaluation-only calibration does not enter model fitting or early stopping. Table~\ref{tab:evidence_recovery} reports representative attention-, instance-, distillation-, severity-, and state-space-based MIL baselines together with all matched label-model controls. DomSet-MIL leads every table metric. Relative to WAN, the next strongest method, it improves Set-mAP by 0.035--0.038 and Hidden-R by 0.089--0.096 across the three controlled tasks. CR-Ordinal has higher dominant-label M-AUC than WAN on all three controlled tasks. WAN leads in Set-mAP and Hidden-R by 0.077--0.090 and 0.140--0.152, respectively. The two evaluations give opposite rankings for these controls.

BRACS permits a separate ROI-level test. DSMIL contributes its native
class-specific instance logits; Flat-CE, WAN, and DomSet-MIL contribute their
class-specific patch scores. The same top-$q$ readout is used for every method.
Each annotated ROI receives the maximum score among patches whose centers fall
inside it, yielding ROI-mAP over ROI labels. Secondary-ROI recall (Sec-ROI-R) is
the fraction of annotated lower-priority ROIs containing at least one of the
slide's top-$q$ patches for that ROI class. ROI annotations remain unavailable
during training. The BRACS columns of Table~\ref{tab:evidence_recovery} report
this comparison; methods without class-specific patch scores are excluded.
DomSet-MIL obtains 0.625 ROI-mAP and 0.500 Sec-ROI-R, versus WAN's 0.594
and 0.451, respectively. These gains reflect improvements in ROI-level evidence
ranking and in the coverage of secondary ROIs by the top-$q$ patches.

\subsection{Ablation study}

We isolate the two factorizations and the pooling operator on BRACS. Replacing
DCL with cross-entropy retains SORR but restores mutually exclusive bag-level
supervision. Replacing SORR with independent class gates retains DCL but removes
residual instance routing. Finally, mean, max, and noisy-OR pooling separately
replace top-$q$ pooling without modifying the gates or bag-level likelihood.
Table~\ref{tab:ablation}
reports both dominant-label performance and secondary-evidence recovery to assess
their trade-offs.

\begin{table}[t]
    \centering
    \caption{Ablation study on BRACS. Best results are bold.}
    \label{tab:ablation}
    \scriptsize
    \setlength{\tabcolsep}{2.2pt}
    \renewcommand{\arraystretch}{1.04}
    \resizebox{\columnwidth}{!}{%
    \begin{tabular}{lccccc}
        \toprule
        Variant & M-AUC & BAcc & UER$\downarrow$ & ROI-mAP & Sec-ROI-R \\
        \midrule
        DomSet-MIL                    & \textbf{0.916} & \textbf{0.690} & \textbf{0.071} & \textbf{0.625} & \textbf{0.500} \\
        \quad DCL $\rightarrow$ CE    & 0.906 & 0.677 & 0.091 & 0.600 & 0.433 \\
        \quad SORR $\rightarrow$ independent gates
                                      & 0.911 & 0.684 & 0.077 & 0.610 & 0.462 \\
        \quad Top-$q$ $\rightarrow$ mean
                                      & 0.903 & 0.670 & 0.089 & 0.596 & 0.442 \\
        \quad Top-$q$ $\rightarrow$ max
                                      & 0.909 & 0.680 & 0.081 & 0.606 & 0.455 \\
        \quad Top-$q$ $\rightarrow$ noisy-OR
                                      & 0.905 & 0.674 & 0.086 & 0.599 & 0.446 \\
        \bottomrule
    \end{tabular}%
    }
\end{table}

Replacing DCL with cross-entropy lowers M-AUC and ROI-mAP by 0.010 and 0.025,
respectively, raises UER by 0.020, and reduces Sec-ROI-R by 0.067. With DCL
retained, replacing SORR with independent gates still reduces Sec-ROI-R by
0.038, supporting residual routing for local evidence recovery. Mean pooling
produces the largest decline in M-AUC and ROI-mAP among the pooling variants.
Max and noisy-OR pooling also reduce both endpoints. Top-$q$ pooling combines
evidence from multiple high-response patches and limits the influence of
low-response regions.

Selective information use is also investigated in multi-agent knowledge
sharing. D3MAS organizes knowledge sharing through
decomposition, deduction, and distribution~\cite{zhang2026d3mas}; graph
decoupling has separately been investigated for agent knowledge
sharing~\cite{zhang2026boosting}.

\section{Conclusion}
\label{sec:conclusion}

We identify a fundamental mismatch in conventional multiclass MIL: a diagnosis
reporting only the most severe lesion censors coexisting lower-severity findings,
which standard training treats as negatives. DomSet-MIL addresses this mismatch
by coupling DCL with SORR and top-$q$ pooling, aligning dominant-label
supervision with severity-ordered patch evidence. Across natural and controlled
settings, it improves dominant-label prediction and latent or localized
secondary-lesion recovery, while ablations support the complementary roles of
censored supervision, residual routing, and sparse pooling. These results show
that modeling the weak-label generation process enables richer pathological
representations from slide-level annotations. Future work will consider
partially ordered, data-dependent, and uncertain label mechanisms, as well as
whether biomedical knowledge-guided prompts, such as those explored in
MeDKCoOp~\cite{wang2025medkcoop}, can support lesion-specific evidence learning.

\clearpage
\section{Compliance with Ethical Standards}

This study used only de-identified or coded pathology images from the released
BRACS, EBRAINS, and TCGA resources; no new participants, specimens, or animals
were involved. All data were used in accordance with the applicable access and
data-use terms.

\bibliographystyle{IEEEbib}
\bibliography{refs}

\end{document}